\documentclass[aps,reprint,superscriptaddress,noeprint,amsmath,amssymb,prb,showkeys,floatfix,nofootinbib]{revtex4-2}

\usepackage{epsfig}
\usepackage{times}
\usepackage{color}
\usepackage{graphicx}
\usepackage{float}
\usepackage{amsmath}
\usepackage[caption=false]{subfig}
\usepackage{appendix}
\usepackage[hidelinks=true,colorlinks=true,linkcolor=blue,citecolor=blue]{hyperref}

\begin{document}

\title{Sampling Latent Material-Property Information From LLM-Derived Embedding Representations}
\author{Luke P. J. Gilligan}

\affiliation{School of Physics, AMBER and CRANN Institute, Trinity College, Dublin 2, Ireland}
\author{Matteo Cobelli}
\affiliation{School of Physics, AMBER and CRANN Institute, Trinity College, Dublin 2, Ireland}

\author{Hasan M. Sayeed}
\affiliation{Materials Science \& Engineering, The University of Utah, Salt Lake City, UT, USA}

\author{Taylor D. Sparks}
\affiliation{Materials Science \& Engineering, The University of Utah, Salt Lake City, UT, USA}

\author{Stefano Sanvito}
\email[Corresponding Author: ]{sanvitos@tcd.ie}
\affiliation{School of Physics, AMBER and CRANN Institute, Trinity College, Dublin 2, Ireland}

\begin{abstract}
Vector embeddings derived from large language models (LLMs) show promise in capturing latent information from the literature. Interestingly, these can be integrated into material embeddings, potentially useful for data-driven predictions of materials properties. We investigate the extent to which LLM-derived vectors capture the desired information and their potential to provide insights into material properties without additional training. Our findings indicate that, although LLMs can be used to generate representations reflecting certain property information, extracting the embeddings requires identifying the optimal contextual clues and appropriate comparators. Despite this restriction, it appears that LLMs still have the potential to be useful in generating meaningful materials-science representations.
\end{abstract}

\maketitle

\section{Introduction}
Recent technological advancements have led to an increase in the throughput of first principles 
simulations of materials. The creation of a variety of repositories of atomic simulations
\cite{Curtarolo_2012,doi:10.1063/1.4812323}, in addition to a steady growth of manually curated 
databases of experimental data \cite{bergerhoff1983inorganic,Allen:an0610,Grazulis2009}, is one of 
the direct consequences of such new capability. This trend, combined with the recent progress in 
Natural Language Processing (NLP) for the automatic generation of databases, extracted from the 
scientific literature~\cite{poly, Gilligan2023}, will rapidly increase the availability of structured 
data that links chemical compounds with their properties. This source of data has proven to 
be a valuable resource for materials informatics when combined with statistical and machine learning (ML)
approaches \cite{PhysRevMaterials.3.104405,doi:10.1126/sciadv.1602241,Isayev2015,doi:10.1021/jacs.8b04704}.

A major challenge in constructing ML models for properties predictions is finding a suitable
representation of the chemical composition, and possibly the structure, of each chemical compound 
the model addresses. A plethora of manually designed descriptors, based on the physical/chemical 
intuition of materials properties, has consequently emerged 
\cite{Ward2016,PhysRevMaterials.3.104405,doi:10.1126/sciadv.1602241}. An alternative approach is to 
generate materials descriptors, usually called materials embeddings, through the NLP interpolation of 
information contained in literature. For example, Tshitoyan {\it et al.}~\cite{Tshitoyan2019} introduced 
Mat2Vec, a Word2Vec model trained on a corpus of texts specific to material science, showing that its 
embeddings perform better than conventional descriptors when used as input features of a shallow neural 
network. In this particular case, the model was able to predict the formation energy of elpasolite 
compounds. Since then, Mat2Vec embeddings have been used in more complex compositional descriptors 
such as CrabNet \cite{Wang2022}. 

Models such as Mat2Vec rely on Word2Vec architectures, which generate so-called static embeddings. 
In these, each word in the model dictionary is assigned to an embedding that is fixed once the model 
is trained. Instead, the recent fast-paced progress in NLP has been led by the transformer architecture 
\cite{vaswani2017attention}. This architecture leverages a self-attention mechanism to generate a 
contextual language representation of the tokens in its dictionary. The contextual awareness of the 
model is a strength in NLP tasks but makes the extraction of the model embedding of an element more 
challenging. In fact, the model representation of a word, for example, 'iron', will depend on the 
context it appears in. To address this problem Bommasani \textit{et al.}~\cite{bommasani-etal-2020-interpreting}
created static embeddings from contextual language models such as BERT \cite{devlin-etal-2019-bert} 
by pooling representations as they appeared in different contexts. This work suggested that for a 
given word one can construct static embeddings containing the information gleaned from the aggregate 
of the contexts available, essentially leveraging the power of language models. A similar philosophy, 
using BERT models, was explored to discover candidate materials for thermoelectric 
applications \cite{bert_reps}.


Given these early successes, it is reasonable to assume that large language models (LLMs) could then 
offer a pathway to manufacture powerful static embeddings, leveraging the immense amount of latent 
information stored in the model parameters. A LLM is a ML model generally based on a decoder transformer
architecture \cite{radford2018improving,radford2019language,brown2020languagemodelsfewshotlearners}. They 
are typically built over billions of model parameters, which encode word relationships of extremely 
extensive corpora of texts. A systematic study on how these static embeddings can be used in materials 
science may give us some clues as to the optimal means of unlocking this colossal amount of information 
in a useful way. There are, however, a number of potential pitfalls, when considering how to construct 
such embeddings for materials science applications.

Firstly, we need to note that LLMs are generally trained autoregressively, meaning that they are 
optimised to identify the most likely word or token to follow a given sequence. This is a potential 
drawback given that the model can only identify the context that informs the next word in a given 
sequence. In other words, due to the causal attention masking, a LLM can only integrate information 
from the context that has appeared before the word of interest in the sequence. For example, in 
the phrase `Iron has a melting temperature of 1,811~K', an autoregressive-model processing of the 
word `iron' will not have access to the melting temperature information that appears later in the 
sentence. This is not the case for a bidirectional model, such as BERT~\cite{bommasani-etal-2020-interpreting},
so that the success of contextualization of BERT embeddings may not transfer to LLMs. In summary, 
as a result of the LLMs' autoregressive nature, a strategy for capturing contextual information from 
static LLMs embeddings may be difficult to conceptualize.

Coupled with this is the relative lack of high-quality contextual information that would be necessary 
for the construction of a suitable embedding using the method outlined by 
Bommasani~\cite{bommasani-etal-2020-interpreting}. In fact, in that case, the total number of contexts 
available for any given word to generate an embedding competitive to that offered by Word2Vec models
was greater than 100,000. This would represent a significant challenge in the materials science domain,
particularly in the case of rare compounds or elements. Such a point is particularly salient considering 
its possible use in materials discovery, which would almost certainly concern domains that are poorly explored in 
the literature.

Another potential issue is that LLMs are tools usually employed for general generative tasks, namely 
for completing word sequences in the general domain. While there is some evidence that LLMs may display 
good performance in domain-specific tasks \cite{thirunavukarasu2023large,openai2024gpt4technicalreport}, 
there exists a trend that fine-tuned small models, trained over narrow highly-focused domains, generally
outperform LLMs over tasks within those domains \cite{ai4science2023impactlargelanguagemodels}. One could 
then potentially come up with a LLM fine-tuning strategy using low-rank adaption (LoRA) \cite{hu2021lora} 
or similar methods. Yet, whether or not the general embeddings encoded in pre-trained LLMs can be used for
materials science tasks, remains a meaningful question.

Thus, this study explores the potential of out-of-the-box LLM embeddings to capture materials information, 
by bringing relationships between composition and intrinsic properties. We will first outline a potential 
minimal strategy for constructing compound-specific embeddings from LLMs, even when there is little to no
information in the training data set of the LLM for a given compound. After this strategy has been outlined, 
a series of explorations will examine the use of these representations for a variety of tasks. Our study 
is meant as an exploratory work and as a means of signposting potential avenues for future research.

\section{Methods}\label{sec:Methods}

All the studies presented in this work have been performed using the output embeddings of open-source 
LLMs. In the main paper, we report the results obtained using the 13-billion (13B) parameters Llama 2 
model from Meta AI \cite{touvron2023llama2openfoundation}. This model has been quantized using 8-bit 
quantization \cite{krishnamoorthi2018quantizing}, such that it could be loaded onto a single Nvidia 
Tesla T4 GPU with 16 GB of VRAM. This Llama 2 model has been chosen after yielding the most promising 
Spearman rank correlation results, when comparing ranking systems with the ground truth rankings, a 
test performed for several other LLMs. In particular, these are Llama 2 with 7B parameters and Llama 3 
\cite{dubey2024llama3herdmodels} with 8B parameters, both trained by Meta AI; the Gemma 1 model 
\cite{gemmateam2024gemmaopenmodelsbased} from Google, with both 2B and 7B parameters; and the Mistral 
AI model \cite{jiang2023mistral7b} with 7B parameters. The results of these comparisons can be found 
in the supplementary information (SI). 

By construction, the language modelling task results in the mapping of the input tokens given to 
the model into high-dimensional vector embeddings in Euclidean space. The geometric relations of 
the resulting embeddings have been shown to reflect the underlying meaning of the language. As such, 
the cosine similarity between words related to each other is higher than that of dissimilar words. 
This can be extended to entire paragraphs or papers and it is the foundation of all modern recommendation 
systems and text retrieval strategies. When assessing the latent material knowledge contained in LLM 
embeddings, we decided to leverage their geometrical properties by evaluating the cosine similarity 
with respect to a relevant word or sentence for the task explored. In the following, we will refer 
to such word or sentence as `query key'.

The first attempt at constructing compound embeddings was performed by taking the embeddings vector 
of the last hidden layer of the model when the chemical formula of a compound was put through a 
forward pass of the model. In this case, the chemical formula was first pre-processed with the suite 
of tools provided by the PyMatGen \cite{ONG2013314} library for standardisation. This standardisation 
involves reordering chemical elements in the compound's string representation so that equivalent formulae 
are represented consistently before being input into the model. The query embeddings used for comparison 
against the compound embeddings via cosine similarity were also obtained by passing the key of interest 
through an equivalent forward pass of the model.

The second attempt at constructing a compound embedding concerns the creation of an individual embedding 
vector for each element in the periodic table, using the same criteria as described above. In this instance, 
the full name of the element is used in place of the symbol of the chemical element (e.g. `iron' against `Fe').
Most importantly, we attempt to enforce the contextualization of the embedding by performing a 
`quasi-contextualisation' step. This consists of adding a biasing term in front of the individual element 
before passing it to the model. For instance, `iron' can be quasi-contextualized in the domain of magnetism 
by passing `ferromagnet iron' to the LLM, instead of just `iron'. The final aggregation of the resulting
embedding is obtained by pooling the embeddings of the individual terms in the resultant phrase. Once 
the embedding vectors for the elements, $\mathbf{v}_{X}$, have been extracted, the compound vector, 
$\mathbf{v}_\mathrm{C}$ can be derived by computing the sum of the elemental embeddings weighted by 
the atomic fraction, $w_{X}$, of that element in the compound,
\begin{equation}
    \mathbf{v}_\mathrm{C} = \sum_{X \in \mathrm{C}} w_{X} \mathbf{v}_{X}\:.
\end{equation}
For example, the feature vector for water ($\mathbf{v}_{\mathrm{H}_2\mathrm{O}}$) is computed as
$\mathbf{v}_{\mathrm{H}_2\mathrm{O}} = 2/3\,\mathbf{v}_\mathrm{H} + 1/3\,\mathbf{v}_\mathrm{O}$, where
$\mathbf{v}_\mathrm{H}$ and $\mathbf{v}_\mathrm{O}$ are the embeddings obtained from a language model for 
water and oxygen, respectively. We call this strategy the composition-averaged elemental embedding.

Each vector $\mathbf{v}_\mathrm{C}$ is then compared with the relevant query key embedding vector
$\mathbf{v}_{q}$, using the cosine similarity, $S_C$. For instance, one would use the $\mathbf{v}_{q}$ 
embedding of `Curie Temperature' to rank magnetic compounds according to their transition temperature. 
The cosine similarity is the dot product between the vectors, scaled to the product of their respective
magnitudes, 
\begin{equation}
S_C=\frac{\mathbf{v}_\mathrm{C}\cdot\mathbf{v}_{q}}{|\mathbf{v}_\mathrm{C}||\mathbf{v}_{q}|},
\end{equation}
a quantity bound to the $[-1,1]$ range. The performance rankings of the resulting compound embeddings are 
thus taken by listing the embeddings according to their $S_C$. 

Finally, the rankings obtained as a result of this methodology are compared with the ground-truth 
performance rankings, taken by listing the compounds in the dataset according to their true property 
values. The comparison between ranking methodologies is performed with a previously mentioned metric, 
known as the Spearman rank correlation coefficient, $\rho$, \cite{ca468a70-0be4-389a-b0b9-5dd1ff52b33f}. 
This measures the degree of similarity between rankings and it is computed as
\begin{equation}
\rho = 1-\frac{6\sum{d_i}^2}{n(n^2-1)}\:,
\end{equation}
where $d_i$ is the difference between the rankings of the data point $i$ using the two methods and 
$n$ is the total number of data points. Two perfectly identical rankings would yield $\rho=1$, perfectly 
inverted rankings give $\rho=-1$ and no correlation returns $\rho=0$.

\section{Results \& Discussion}

Our first test is taken outside the realm of materials science and arguably in a domain with a 
greater abundance of online information. We use the final-layer representation of all the countries 
in the world and compute their cosine similarity with the representation derived from a variety of 
keywords that are generally associated with indicators of a country's economic performance. The 
cosine-similarity rankings of these countries are then compared with the rankings based on the 
gross domestic product (GDP) of the country for the year 2022, as taken from the World Bank data
\cite{worldbank_gdp}. The first test is performed by comparison with the keyword query phrase 
`gross domestic product' (the selected $\mathbf{v}_{q}$ vector).
\begin{figure}[ht!]
    \centering
    \includegraphics[width=0.8\linewidth]{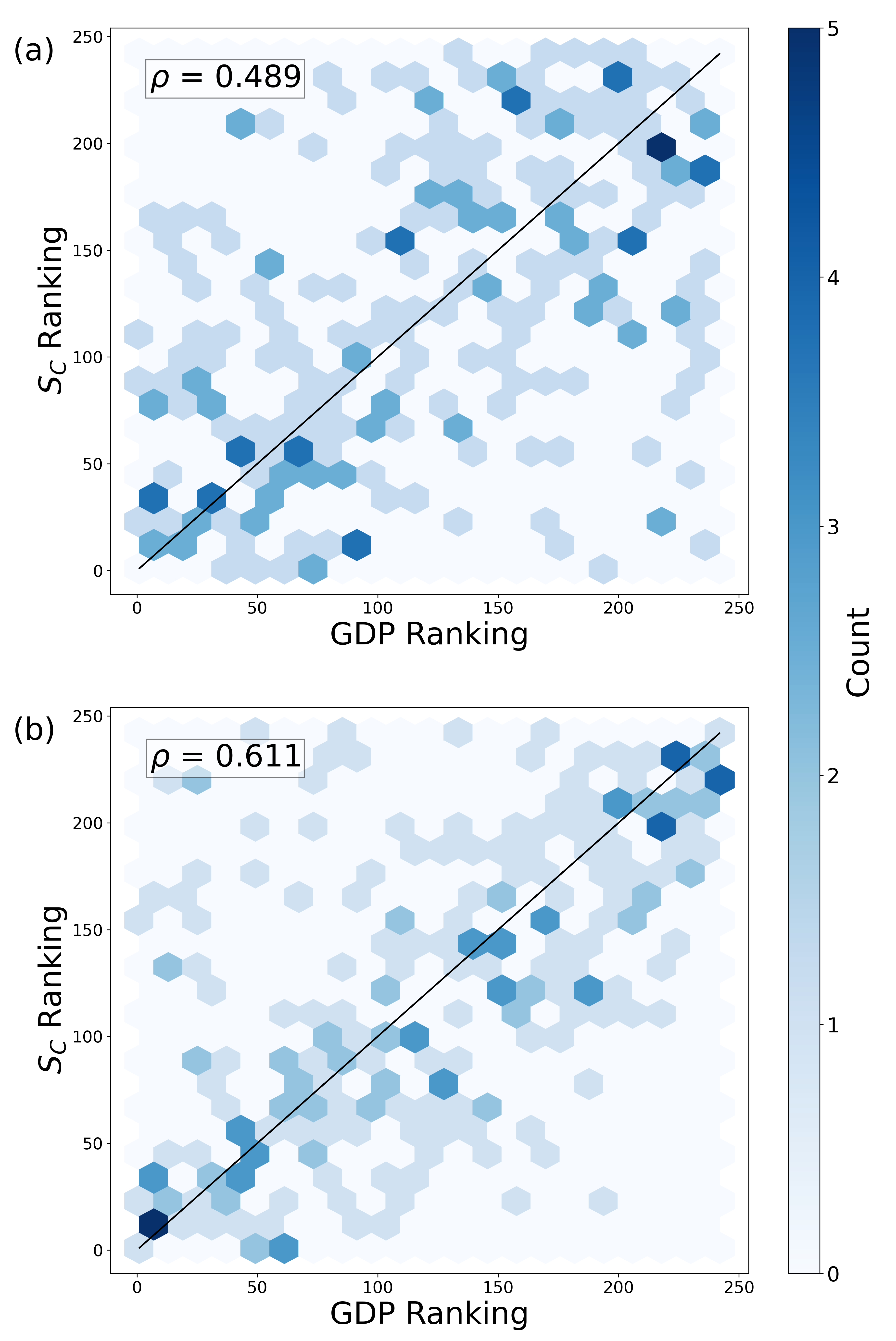}
    \caption{Parity plots comparing the World Bank 2022 GDP ranking (`GDP ranking') with the country 
    cosine similarity against the string `gross domestic product'. Here the embeddings are derived from 
    the final-layer LLM representation: \textbf{(a)} without any context and \textbf{(b)} by providing 
    the contextual phrase `economy of' before the country name. Both rankings are compiled using the 
    largest Llama 2 model (13B parameters). The colours encode the number of countries presenting that 
    particular ranking.}
    \label{fig:gdp}
\end{figure}

We observe that there is already a strong positive correlation between the rankings based on the 
GDP of a country and the cosine similarity of the word representation obtained from the largest LLM 
available, Llama 2 (13B parameters). Such correlation is obtained even without the provision of any 
context, as depicted in the parity plot of Fig.~\ref{fig:gdp} (a), showing the cosine similarity of 
each country against the actual 2022 GDP ranking. This plot shows a Spearman rank correlation 
$\rho = 0.489$, a result that provides an incentive for the use of pre-trained LLMs to indicate 
metrics that can be associated with derived word representations.

Interestingly, simply introducing a short prefix before the country serves to enhance the predictive 
power of the representation, as evidenced by the improved ranking performance visible in 
Fig.~\ref{fig:gdp} (b). In this case, the embedding of a country $\langle$country$\rangle$ is extracted 
from querying the LLM with the sentence `economy of $\langle$country$\rangle$', and by mean-pooling 
the resulting embedding for $\langle$country$\rangle$ ($\langle$country$\rangle=$ Ireland, Italy, Spain, 
etc.). This simple contextualization improves the Spearman rank correlation from $\rho=0.489$ to 
$\rho=0.611$. Thus, a potential avenue for tuning the quality of the predictive performance of the 
methodology is potentially uncovered by a suitable choice of the context provided to the token of 
interest.
\begin{figure}[ht!]
    \centering
    \includegraphics[width=1\linewidth]{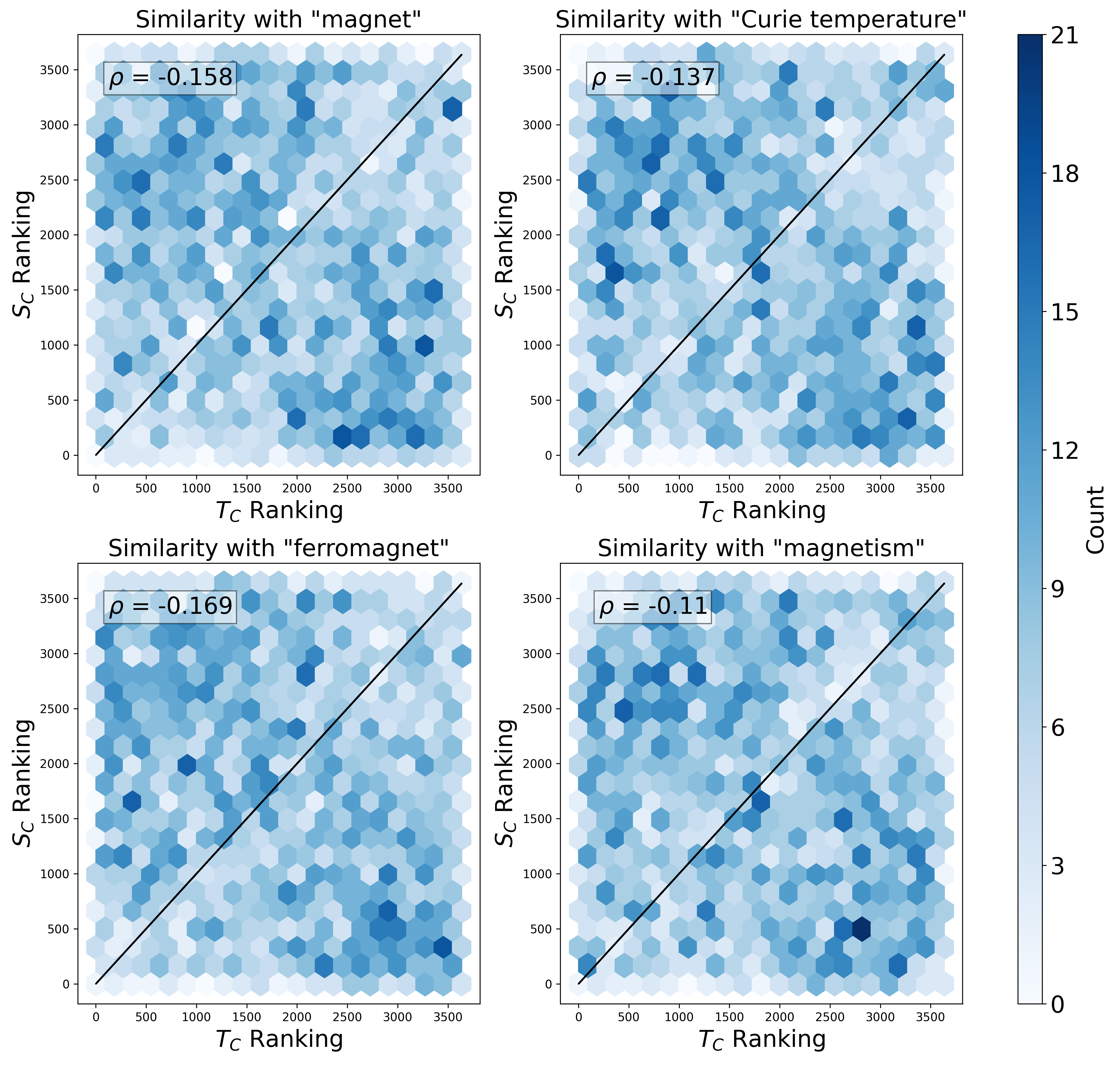}
    \caption{Parity plots comparing the ground-truth Curie-temperature ranking with the ranking based on the cosine similarity of the embedding vectors with different magnetic keywords (reported above each graph). In this case, the chemical formulae are directly embedded by the LLM. All rankings are compiled using the largest Llama 2 model (13B parameters). The Spearman rank correlation of each plot is reported in the legends. The colours encode the number of compounds presenting that particular $T_\mathrm{C}$ ranking.}
    \label{fig:magnet_whole}
\end{figure}

Let us now apply the same strategy in the domain of materials science. In this case, we consider 
a database of Curie temperatures, $T_\mathrm{C}$, of ferromagnets, obtained by aggregating manually 
curated datasets available in the literature. These are the databases of Nelson {\it et al.}~\cite{PhysRevMaterials.3.104405}, containing data from \textit{AtomWork}~\cite{Yamazaki2011}, 
\textit{Springer Materials}~\cite{Connolly2012}, the \textit{Handbook of Magnetic Materials}~\cite{Handbook} 
and the book {\it Magnetism and Magnetic Materials}~\cite{Coey}, and the database of Byland {\it et
al.}~\cite{byland}, which mainly focuses on Co- containing compounds. In total, we can rank 3,638 
unique compounds according to their Curie temperature, a ranking that represents our ground truth. 
This is compared with the cosine-similarity ranking first obtained by simply passing the chemical 
formula of a compound (instead of the composition-averaged elemental embedding) through the LLM without 
a contextualization term. The results of this first exercise are presented in Fig.~\ref{fig:magnet_whole} 
for a range of query keys that, by intuition, should correlate with the performance of a magnetic 
material. These are `magnet', `Curie temperature', `ferromagnet' and `magnetism'.

\begin{figure}[ht!]
    \centering
    \includegraphics[width=1\linewidth]{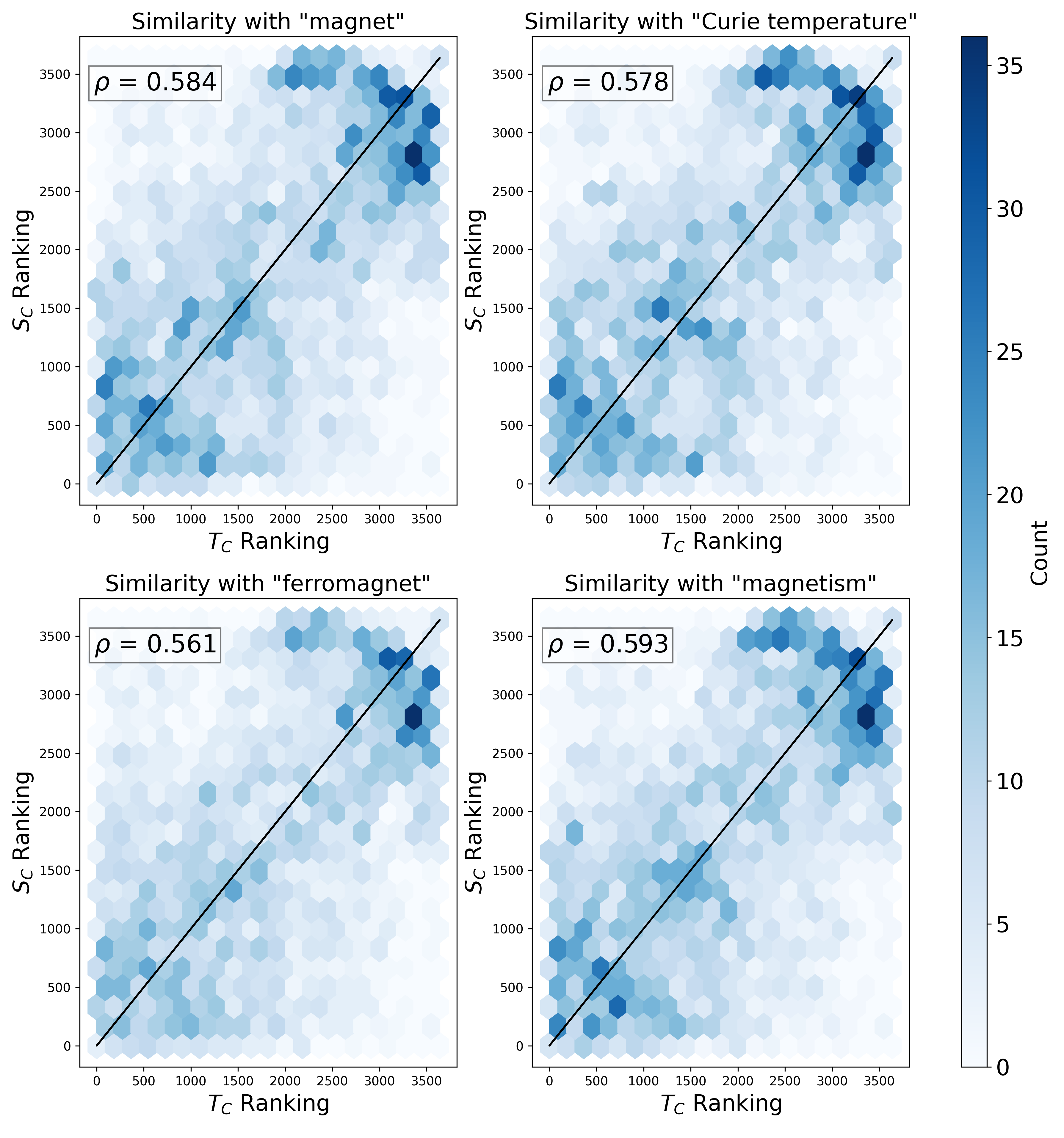}
    \caption{Parity plots comparing the ground-truth $T_\mathrm{C}$ ranking with the ranking based on the cosine similarity of the embedding vectors with different magnetic keywords (reported above each graph). In this case, each compound is embedded through the composition-averaged elemental embedding, having `ferromagnet' as a contextualization term. All rankings are compiled using the largest Llama 2 model (13B parameters). The Spearman rank correlation of each plot is reported in the legends. The colours encode the number of compounds presenting that particular $T_\mathrm{C}$ ranking.}
    \label{fig:magnet_qc}
\end{figure}

From the figure, it is clear that the simple injection of chemical formulae into a LLM is not a 
valid methodology for building a training-free method for ranking materials based on their propensity 
to ferromagnetism. In fact, none of the query keys yield a positive Spearman rank 
correlation coefficient, which is close to zero in all cases, showing that the rankings are not 
correlated. This is pretty evident from the parity plots, which display rather uniform distributions 
of points. Thus, a new philosophy must be explored. As such we now consider the composition-weighted 
elemental representation of the chemical formulas (see Section~\ref{sec:Methods}). In particular, 
we extract the embedding of the elements by contextualizing the search with the word `ferromagnet'. 
The results of this new exercise are reported in Fig.~\ref{fig:magnet_qc} for rankings obtained 
against the same query keys as before: `magnet', `Curie temperature', `ferromagnet' and `magnetism'.

As it can be seen in the figure, this different embedding strategy leads to far superior 
performance in ranking compounds according to their $T_\mathrm{C}$, in particular at the 
two extremes of the ranking (low and high temperatures). The two most successful query 
keys are `magnet' and `magnetism', but in general for all the ones investigated, we find 
a Spearman rank correlation above 0.5. This suggests that LLMs without any domain-specific 
tuning already have some capability to relate compounds's $T_\mathrm{C}$ of magnets relative 
to each other

This significant improvement in the predictive performance of LLM is welcome and entirely 
unexpected. One can then wonder about the origin of such success, namely whether it is related 
to some element of context-awareness of the LLM in the domain of magnetism or to the structure 
of our composition-averaged elemental embedding strategy. This second option could be likely in 
the case of ferromagnets, which present a high degree of correlation between their Curie temperature 
and their composition, in particular with the presence and concentration of some key elements such 
as iron, cobalt and rare earth elements~\cite{PhysRevMaterials.3.104405,Belot2023}.

\begin{figure}[ht!]
    \centering
    \includegraphics[width=\linewidth]{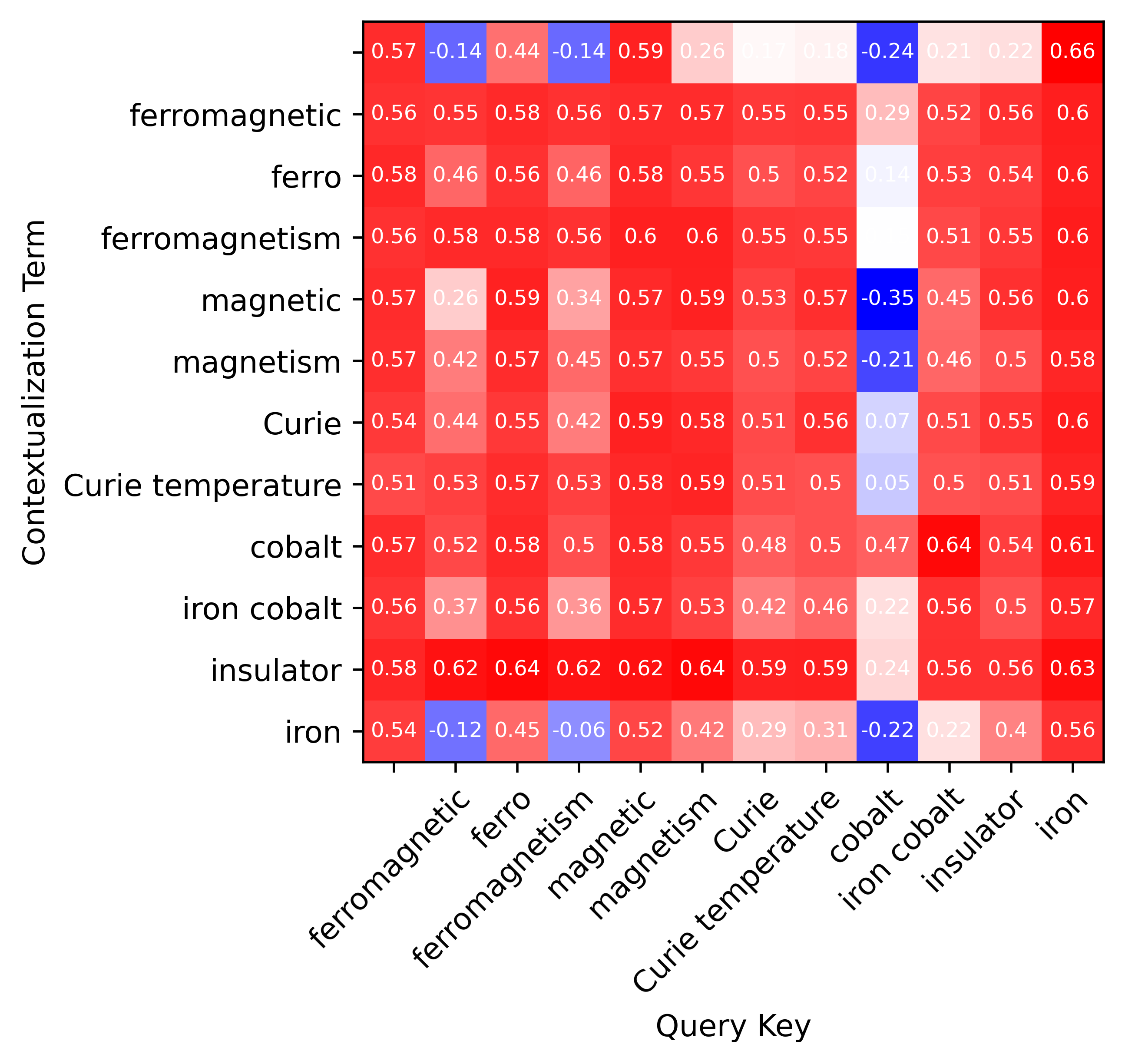}
    \caption{A heat map of the Spearman rank correlation coefficient, $\rho$, for different choices of contextualization terms and query keys. This is computed against the ground truth Curie-temperature database. The first row corresponds to composition-averaged elemental embedding in which no contextualization term was introduced, while in the first column, the query key is an empty string. The systematically best-performing query key is `iron'.}
    \label{fig:magnet_heatmap}
\end{figure}

We can test the two different hypotheses by comparing the Spearman rank correlation for a variety 
of different contextualization terms and query keys. In particular, we explore 11 different terms 
and query keys, comprising generic terms related to magnetism (e.g. `magnetism', `ferromagnetic', 
`Curie Temperature', etc.) and the name of a few elements highly present in ferromagnets (e.g. 
`iron' and `cobalt'). We then compute the Spearman rank correlation for each of the combinations,
with our results being presented in Fig.~\ref{fig:magnet_heatmap}.

In general, we find that many contextualization terms perform equally well against most of the 
query keys. For instance, using `ferromagnetic', `Curie temperature' or even just `Curie' returns 
us a Spearman rank correlation just above 0.5 regardless of the query key. The only exception is 
when querying with `cobalt', which gives us rather uncorrelated rankings for all the embeddings 
tested, indicating that the performance of a ferromagnet (its position in a $T_\mathrm{C}$ ranking 
system) cannot be assessed through its LLM similarity with the element cobalt. Interestingly, the 
opposite is not true, namely contextualizing the elemental embedding with `cobalt' seems to generate 
vectors, which then rank well according to many of the query keys. Intriguingly, the situation with 
`iron' is exactly the opposite, namely the word `iron' provides modest contextualization (the 
Spearman rank correlation is modest for all query keys), but the same term appears to be the most 
performing in producing a cosine-similarity ranking of compounds. This may be related to an intrinsic 
correlation in materials science, between a compound's iron content and its 
$T_\mathrm{C}$~\cite{PhysRevMaterials.3.104405}. Another interesting, and somehow expected result, 
is that the contextualization with terms poorly related to magnetism, such as `insulator', or with 
an empty string usually generates embeddings resulting in a weak correlation. More surprising is 
that the same is found for both `magnetic' and `magnetism' against magnetism-related queries. Finally, 
it is worth noting that the ranking computed according to the similarity with an empty key (first 
column) seems to be also rather good, a fact that we may consider accidental. 

The results presented here for the Llama 2 model (13B parameters) are somehow mirrored by other
LLMs (see figures 1, 4, 7, 10 and 13 in the SI), although we notice significant scattering in the 
actual data. It is also interesting to note that the performance of a given model in this ranking 
task does not seem to be tightly correlated to the size of the model. For instance, we find the 
heat map of the Spearman rank correlation coefficients of Gemma 2B-parameters (Fig.~1 in the SI) 
to be more similar to the one presented here for Llama 2 (13B parameters), than that obtained with 
Gemma 7B-parameters (Fig.~4 in the SI). For this latter one, for instance, constructing the ranking
with `iron' as query key does not lead to positive Spearman rank correlation coefficients, except 
for the case where the contextualization is obtained with `ferro' and `ferromagnetic'. Such differences 
in the various LLMs' performance may boil down to the different data sets used for their training.
 
As noted above, there is a tendency for the $T_\mathrm{C}$ of ferromagnets to be correlated with 
the iron content, a fact that may obfuscate our assessment of the natural ability of the LLM to 
contextualize the embedding of elements in the magnetism domain. Thus, a second set of tests is 
performed against two material properties for which there is less correlation between the property 
and the atomic fraction of a particular element. In particular, we consider possible rankings 
according to properties related to the thermoelectric performance, namely the power factor and the 
band gap, where the ground-truth values are contained in the datasets of Ricci {\it et al.}~\cite{ricci2017ab}
and Zhuo {\it et al.}~\cite{BandGapData}, respectively. These properties are significantly less 
compositionally dependent, or better they depend less on the presence of a single element, with 
crystal structure, hybridisation, and other physical/chemical characteristics all contributing. 
Note, however, that some caution should be taken when looking at these quantities. In fact, in 
general, the power factor depends on the doping type and level of a particular compound, while 
there are multiple definitions of bandgap (spectroscopical, transport, etc.), resulting in different 
values. This suggests that the ground-truth rankings may be affected by some `noise' intrinsic to 
the definition of the quantity of interest.

\begin{figure}[ht!]
    \centering
    \includegraphics[width=1.0\linewidth]{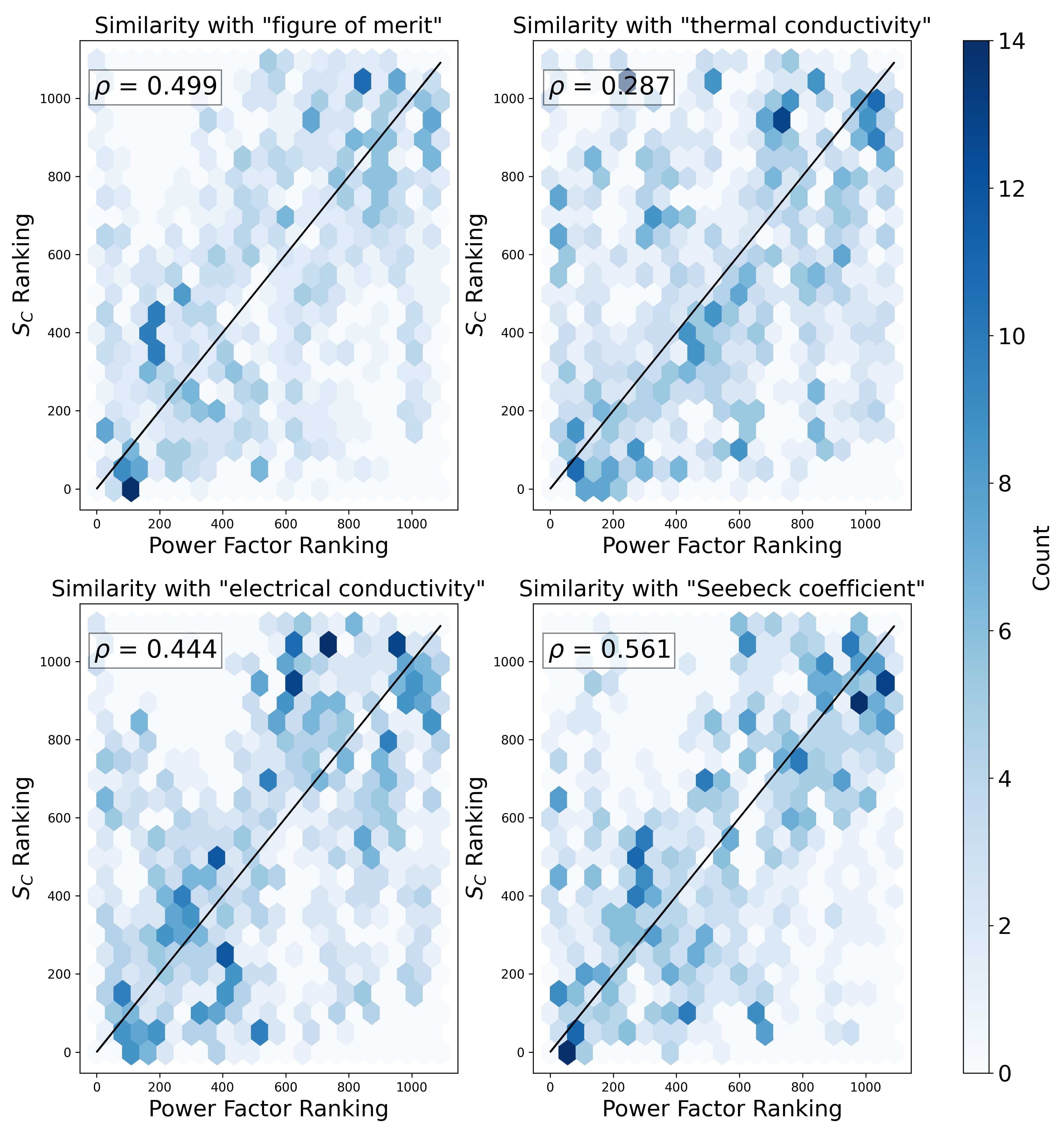}
    \caption{Parity plots comparing the ground-truth thermoelectric power factor with the ranking based on the cosine similarity of the embedding vectors with different thermoelectric-related query keys (reported above each graph). In this case, each compound is embedded through the composition-averaged elemental embedding, having `thermoelectric' as the contextualization key. All rankings are compiled using the largest Llama 2 model (13B parameters). The Spearman rank correlation of each plot is reported in the legends. The colours encode the number of compounds presenting that particular power-factor ranking.}
    \label{fig:thermoelectric_qc}
\end{figure}

Thus, the next experiment consists of reproducing the ground-truth power-factor ranking by using 
a composition-averaged elemental embedding contextualized with the term `thermoelectric' and queried 
with the keys `figure of merit', `thermal conductivity', `electrical conductivity' and `Seebeck 
coefficient'. As it can be noted from the parity plots of Fig.~\ref{fig:thermoelectric_qc}, once 
again there is a strong correlation between the representation similarities of the compositional 
embeddings and the similarity of the query key in most cases. The query key with the weakest 
correlation in this instance is `thermal conductivity', while that with the strongest is `Seebeck 
coefficient'. This is a promising result since the power factor does not have a dependency on 
thermal conductivity, while it does have a quadratic dependence on the Seeback coefficient, a fact 
which could account for the relative strengths of the correlations associated with these two ranking 
query keys. The co-occurrence of thermoelectric material mentioned in literature with discussions 
about thermal conductivity, an important quantity for the thermoelectric figure of merit $zT$, could 
go some way towards accounting for the weak correlation observed.

The degree of ranking correlation and the strength of the relationship between the query key 
and the thermoelectric power factor are likely not to be directly represented within the model. 
Therefore, this should not be interpreted as evidence that the LLM has any inherent mathematical 
understanding of the relationship. It might, however, betray some sort of statistical awareness of 
the relationship based on the context in which the element may appear in literature used in the 
training of the LLM. Elements generally associated with high thermoelectric performance may 
co-occur in the literature with mentions of the Seebeck coefficient, whereas elements which often 
feature in discussions related to thermal conductivity will likely not correlate to high thermoelectric
performance.

\begin{figure}[ht!]
    \centering
    \includegraphics[width=\linewidth]{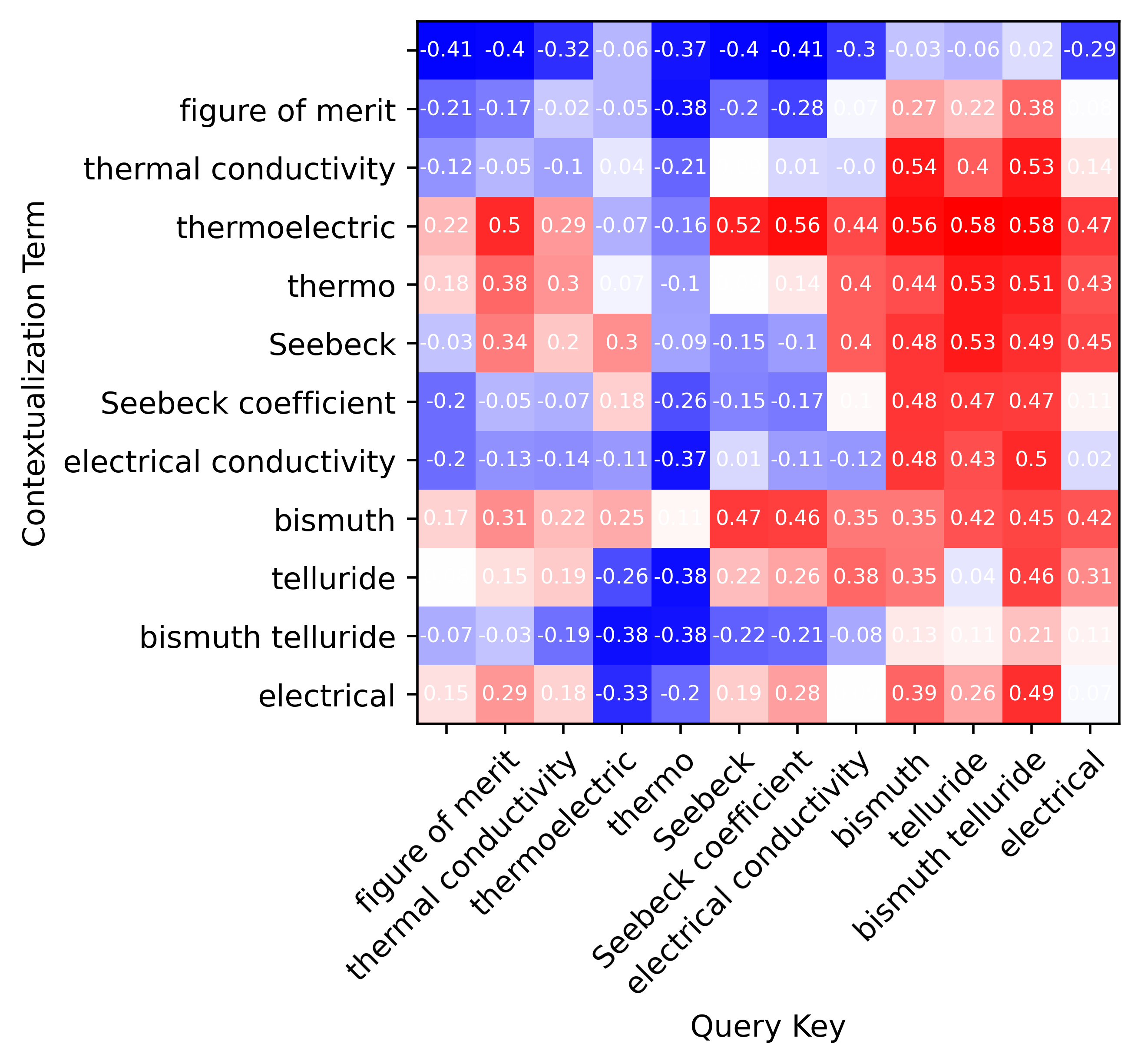}
    \caption{A heat map of the Spearman rank correlation coefficient, $\rho$, for different choices of contextualization terms and query keys. This is computed against the ground truth thermoelectric power factor database. The first row corresponds to composition-averaged elemental embedding in which no contextualization term was introduced, while in the first column, the query key is an empty string.}
    \label{fig:thermoelectric_heatmap}
\end{figure}

Further insights can be gained by looking at the heat-map plots of the Spearman rank correlation 
coefficient, cross-relating embedding contextualization terms and query keys, presented in Fig.~\ref{fig:thermoelectric_heatmap}. Again we consider domain-specific terms, such as `thermoelectric', 
`figure of merit', etc., together with material-specific ones (e.g. `bismuth') and with the empty key. 
In this case, we find a much less successful correlation than that observed for the magnetic-materials 
example. Certainly, the most successful contextualization term seems to be `thermoelectric', which 
results in a positive ranking correlation against almost all query keys. Then, with the exception of 
`thermo' and `bismuth', which still show positive Spearman rank correlation coefficients, all the other 
contextualization terms perform rather poorly, typically at par with embeddings obtained with no
contextualization. Turning the attention to the query keys, it is quite clear that `bismuth', `telluride' 
and `bismuth telluride' appears to be the most proficient at establishing a cosine similarity ranking 
similar to that obtained by looking at the power factor. This time there is no known relation between 
the quantity of interest, the power factor, and the composition fraction of an element in a compound. 
However, it is also true that bismuth- and tellurium-based compounds are typically among the good 
thermoelectric materials. As such, we may dare to formulate the hypothesis that the LLM has captured 
some semantic similarity between the constructed compound representation and the representation 
associated with the compound best known as a high-performing thermoelectric material.
Again, qualitatively similar results are obtained for other LLMs (see figures 2, 5, 8, 11 and
14 in the SI).

It is also worth remarking that in this instance, the effect of biasing the elemental representation 
used to construct the compound representation towards the domain of thermoelectricity is clear, with 
this context term systematically offering the most similar rankings to the ground truth. Such a result 
supports the finding derived from the study involving the GDP and does imply that a similar contextualisation, 
in fact, somewhat improves ranking performance in certain domains.

\begin{figure}[ht!]
    \centering
    \includegraphics[width=\linewidth]{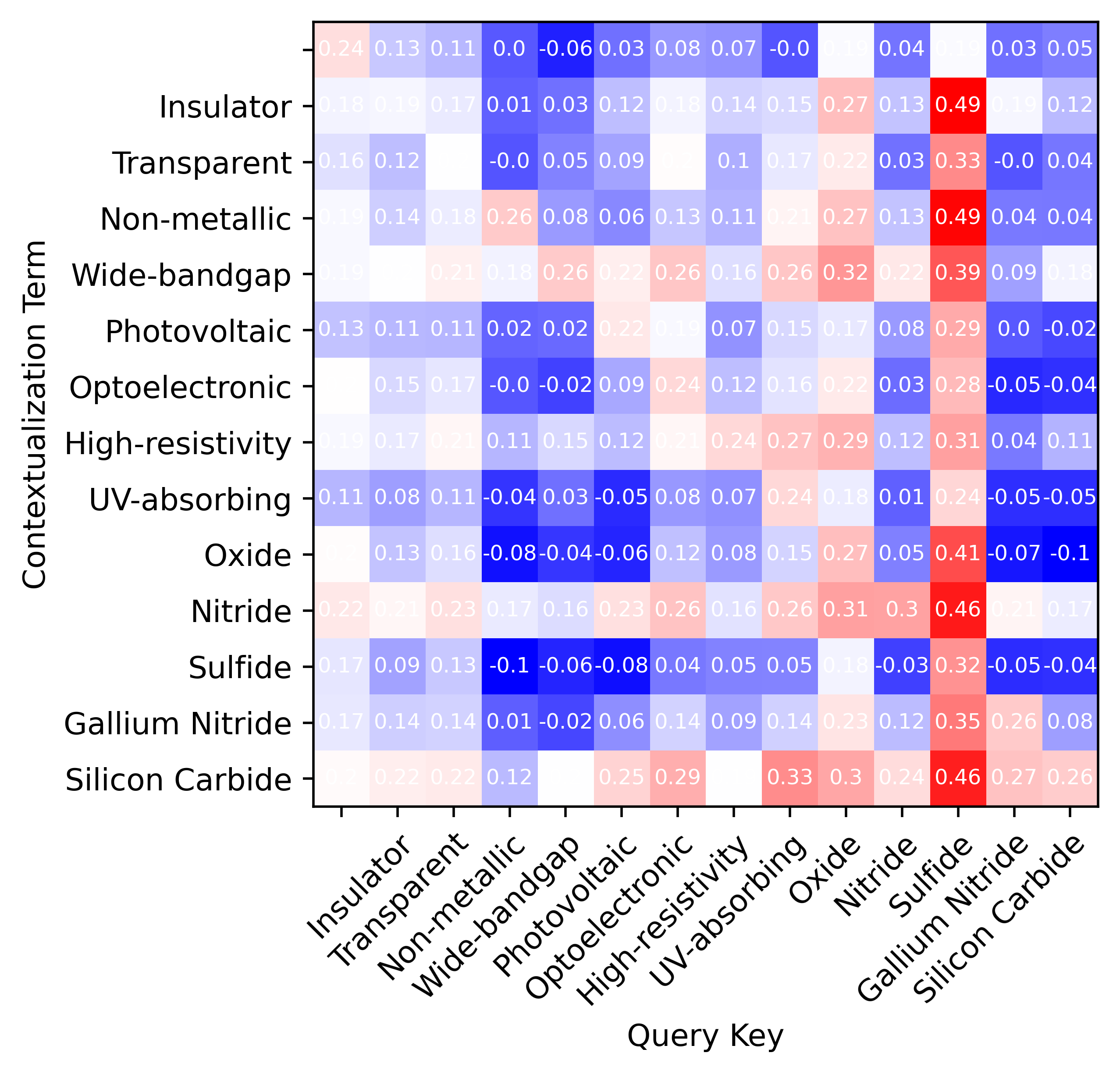}
    \caption{A heat map of the Spearman rank correlation coefficient, $\rho$, for different choices of contextualization terms and query keys. This is computed against the ground truth band gap database. The first row corresponds to composition-averaged elemental embedding in which no contextualization term was introduced, while in the first column, the query key is an empty string.}
    \label{fig:band_heat}
\end{figure}

The final property studied using this methodology is the electronic band gap. This typically has 
a complex dependence on the chemical and structural characteristics of a compound and should be poorly 
correlated to the abundance of a particular element. Thus, in order for any relationship with the ground 
truth to emerge in this ranking, the embedding of a compound should have incorporated some contextual 
information. This, however, seems not to be the case. In fact, the Spearman-rank correlation coefficient
heat map of Fig.~\ref{fig:band_heat} shows little to no correlation for most choices of contextualization 
terms and query keys. Once again in this instance, the bias of the computed representation towards a 
high-performing material in the domain of interest does seem to improve the ranking performance of the methodology, with the term `sulfide' as query key systematically offering the best performance. Other query 
keys do not offer the same indication of a systematic improvement, however, when the correct 
contextualization term is found to construct the representation, the best performing involve the propensity 
of the material not to conduct electricity, i.e.~`insulator' and `non-metallic', or are again related to a 
likely high-performing materials such as `nitride' or `silicon carbide'. It is, however, difficult to 
establish if there is anything systematic that would allow for the construction of a reasonably 
high-performance ranking system out of the box, without optimizing both the query and contextualization 
term over a held-out set of labelled data. To some degree, this issue presents many similarities to the 
problem of prompt engineering. The outputs of LLMs are dependent on the input prompt passed to them, 
and small alterations to the prompt can sometimes lead to significant changes in the corresponding 
output. In this sense, a more robust term-engineering strategy is required to increase the reliability 
of property-based ranking using LLM embeddings. The results obtained with other LLMs are shown in 
Figures 3, 6, 9, 12 and 15 in the SI.

%

Thus, in conclusion, there is some evidence that the latent information contained within an LLM 
could help the construction of some form of valid representation for materials science applications. 
The main issue identified in this work is the difficulty in establishing a scheme, which can systematically
unlock such information in a useful and consistent way. LLMs out-of-the-box will rarely contain sufficient
information to construct an adequate representation in a decontextualised setting, but a full aggregation 
of contextual information may not be entirely necessary if a contextualisation term biases the representation
sufficiently towards the domain of interest.

\section{Conclusion}

In summary, we have investigated the ability of LLMs to construct material-science embeddings, which 
transfer some of the contextual knowledge built into the model to the embeddings themselves. This may 
potentially replace the formulation of physics- and chemistry-informed descriptors for machine-learning 
models. In particular, we have considered the field of magnetism and that of thermoelectricity. Our main 
testing strategy has been that of extracting embeddings from the terminal layers of the LLMs, after 
having passed to the model either the chemical formula or the name of an element, together with a 
contextualization term. The embeddings of the chemical formulas can be used directly, while in the case 
of elements we formulate a composition-averaged elemental descriptor, which describes any given compound. 
Then, the resulting embeddings are finally ranked according to the cosine similarity with some query key, 
which describes the property of interest or what is believed to be a reasonable context. The resulting 
ranking system is compared to the ground-truth ranking through the Spearman rank correlation coefficient.

We have found that the direct embedding of chemical formulas does not produce rankings of quality. In 
contrast, the composition-averaged elemental embeddings seem to perform reasonably well in particular 
when extracted with a relevant contextualization term. This suggests some ability of the LLM to 
contextualize words in the domain of interest. However, we have also found that the best results are 
obtained for quantities, such as the ferromagnetic Curie temperature, which strongly correlate with the 
presence and fraction of a particular element in a compound. This is the case of iron in ferromagnets. 
When such a bias is removed, as in the case of the band gap, the embeddings seem to be significantly 
less performing and they usually correlate only with query keys that identify a proficient material 
according to the property of interest (e.g. bismuth telluride in the case of a thermoelectric power 
factor). Finally, it is also to be noted that different LLMs seem to perform quantitatively and sometimes
qualitatively differently from each other, possibly reflecting the content of the datasets the models 
have been trained on.

All in all, we can then conclude that LLMs are likely not to be a valuable means of 
gaining an estimate of relative material property ranking immediately, at least when they are taken without any training 
or optimisation step. However, a common sense choice of contextualisation term and query key may be 
useful in certain contexts. This work serves to highlight the potential strengths and drawbacks of LLMs 
for constructing valuable materials representations for data-driven discovery.

\section*{Data Availability}
The dataset of Curie temperature is available under request from the authors. The datasets related to
thermoelectric properties and bandgaps are available at \cite{ricci2017ab} and \cite{BandGapData}, 
respectively.

\section*{Code Availability}
The scripts used to generate the compound representations can be found at \url{https://github.com/StefanoSanvitoGroup/LLM-Material-Embeddings/}.

\section*{Acknowledgments}
This work has been sponsored by the Irish Research Council, through the Advance Laureate Award, IRCLA/2019/127. Additional support has been provided by the Science Foundation Ireland AMBER Centre (12/RC/2278$_-$P2). We acknowledge the DJEI/DES/SFI/HEA Irish Centre for High-End Computing (ICHEC) and Trinity Centre for High-Performance Computing (TCHPC) for the provision of computational resources.

\section*{Author Contributions}
The project was designed by LPJG, MC and SS. All calculations and analyses have been performed by LPJG, MC and HMS. The first draft of the manuscript has been prepared by LPJG and MC, and it was reviewed by all the authors. TDS and SS provided the supervision of the project.

\section*{Competing Interests}

The authors declare no competing interests.

\bibliographystyle{naturemag}
\bibliography{references-new}

\pagebreak

\listoffigures
\vfill
\end{document}